\documentclass[sigconf]{acmart}
\AtBeginDocument{%
  \providecommand\BibTeX{{%
    \normalfont B\kern-0.5em{\scshape i\kern-0.25em b}\kern-0.8em\TeX}}}

\setcopyright{acmcopyright}
\copyrightyear{2022}
\acmYear{2022}
\setcopyright{usgovmixed}\acmConference[GECCO '22 Companion]{Genetic and Evolutionary Computation Conference Companion}{July 9--13, 2022}{Boston, MA, USA}
\acmBooktitle{Genetic and Evolutionary Computation Conference Companion (GECCO '22 Companion), July 9--13, 2022, Boston, MA, USA}
\acmPrice{15.00}
\acmDOI{10.1145/3520304.3534048}
\acmISBN{978-1-4503-9268-6/22/07}

\usepackage{booktabs}
\usepackage{algorithm}
\usepackage{algorithmic}
\usepackage{caption}
\usepackage{subcaption}
\begin{document}

\title{Evolving SimGANs to Improve Abnormal Electrocardiogram Classification}

\author{Gabriel Wang, Anish Thite, Rodd Talebi, Anthony D'Achille, Alex Mussa, and Jason Zutty}

\affiliation{
  \institution{Georgia Tech Research Institute}
  \country{United States}
}

\begin{abstract}
Machine Learning models are used in a wide variety of domains. 
However, machine learning methods often require a large amount of data in order to be successful.
This is especially troublesome in domains where collecting real-world data is difficult and/or expensive.
Data simulators do exist for many of these domains, but they do not sufficiently reflect the real world data due to factors such as a lack of real-world noise.
Recently generative adversarial networks (GANs) have been modified to refine simulated image data into data that better fits the real world distribution, using the SimGAN method. \cite{SimGAN}.
While evolutionary computing has been used for GAN evolution, there are currently no frameworks that can evolve a SimGAN.
In this paper we (1) extend the SimGAN method to refine one-dimensional data, (2) modify Easy Cartesian Genetic Programming (ezCGP), an evolutionary computing framework, to create SimGANs that more accurately refine simulated data, and (3) create new feature-based quantitative metrics to evaluate refined data.
We also use our framework to augment an electrocardiogram (ECG) dataset, a domain that suffers from the issues previously mentioned. 
In particular, while healthy ECGs can be simulated there are no current simulators of abnormal ECGs.
We show that by using an evolved SimGAN to refine simulated healthy ECG data to mimic real-world abnormal ECGs, we can improve the accuracy of abnormal ECG classifiers.
\end{abstract}



\maketitle

\section{Introduction}
For certain problems and domains, collecting real-world data can be difficult and expensive.
Simulations are used to mimic real-world stimuli and can provide a clear, valid, easily reproducible digital representation of a system, often founded on principles of mathematics and physics.
However, oftentimes simulations may not be able to capture the full complexity of real-world data due to poor assumptions or bias built into the model. This leads to a gap between synthetic and real data distributions, and ultimately a machine learning solution biased towards the larger synthetic dataset.
This is a common problem in the medical industry: if we want to study heart disease, collecting electrocardiogram (ECG) heartbeat signals from real subjects can be a time-consuming and tedious process, as medical data needs to be properly classified and anonymized, often requiring a trained professional.
There are many existing simulations that mimic a healthy heartbeat \citep{PhysioBank, simple_ecg_sim, advanced_ecg_sim, NeuroKit2}, but they poorly capture the noise seen in the real data.
The complementary unhealthy heartbeat data has yet to be simulated consistently well.

Generative Adversarial Networks (GANs) \citep{GAN} have seen great success in generating synthetic data samples of different types such as images, speech, text, etc., by having two neural networks compete against one another.
Furthermore, researchers have modified GANs to be used in combination with simulators to take advantage of simulated and unsupervised learning, in the SimGAN learning method \citep{SimGAN}.
Rather than having the GAN start from random noise as its input, as is traditionally done with GANs, a simulator's output is fed to the GAN's input. This change in procedure allows the GAN to take advantage of the simulator's annotations while producing more realistic samples compared to the original simulator.
However, GANs are predominantly used for image generation or 2D data types, and 1D GAN data synthesis has not seen as many real world applications.
Despite the successes of these models and methods, deep generative models are still difficult to train and evaluate. This difficulty stems from generative models' lack of objective truth. Without truth, there are not always easy quantitative metrics for determining how realistic the models' outputs are.
Creating a successful network architecture and finding the optimal hyperparameters for it are also challenges, as it often requires a researcher to manually tune the models through trial and error.
On top of that, a variety of loss functions \citep{GAN, WGAN}, network architectures \citep{DCGAN, styleGAN}, and training techniques \citep{mbd_feature_matching_loss, WGAN-GP, DRAGAN} have all been proposed, so it is difficult to select which ones would work best for a specific domain.
 
While evolutionary computation (EC) has been utilized before to help optimize and search for successful GAN architectures \citep{E-GAN} and loss functions \citep{TaylorGAN}, we could not find an evolutionary system that can perform neural architecture search for a SimGAN, adjust custom loss functions, train and evaluate models across multiple objectives, and optimize hyperparameters simultaneously while being customizable to fit any data problem, including 1D data.

Thus, in this paper, we pose the following research questions:
\begin{itemize}
\item \textbf{RQ1:} How effective are SimGANs for generating realistic 1D signals from simulated inputs? Are SimGANs able to shift the synthetic data distribution closer to the real data distribution? Are they applicable to real world problems like ECG heartbeat synthesis?
\item \textbf{RQ2:} What are examples of effective quantitative metrics for evaluating SimGAN outputs during and after training? 
\item \textbf{RQ3:} How can we utilize evolutionary computation to search for novel SimGAN architectures and optimize existing models? Can we automate this process so that training, evaluation, optimization, and selection of models can be done simultaneously while being flexible enough to be applied to a multitude of problems? 
\end{itemize}

The main contributions of this paper are:
\begin{enumerate}
    \item Improvements to the SimGAN learning approach and architecture for 1D data
    \item New feature-based quantitative metrics for evaluating generated outputs
    \item An open source software implementation of ezCGP, an evolutionary computation framework for optimizing complex machine learning pipelines
    \item Examples of SimGAN outputs mimicking abnormal heart conditions from the MIT-BIH arrhythmia database \citep{MIT-BIH} and the effects of using the outputs for heartbeat classification
\end{enumerate}

The rest of the paper is organized as follows; Section \ref{Background} presents a brief overview of important concepts and related work. 
Section \ref{Dataset} describes the dataset and simulators we used. 
Section \ref{Methods} describes our SimGAN improvements and ezCGP's evolutionary process. 
The results and discussion are in Section \ref{ResultsAndDiscussion}. Finally, conclusions are drawn and future work is outlined in Section \ref{ConclusionAndFutureWork}.

\section{Background}
\label{Background}
This section introduces some basic concepts in GAN training, the SimGAN learning method, our ezCGP framework, and summarizes relevant studies related to this research.
\subsection{General GAN training}
GANs \citep{GAN} typically train two deep neural networks (DNNs): a generator ($G$), which generates synthetic samples, and a discriminator ($D$), which predicts if the synthetic samples are real or fake.
These two networks are placed in an adversarial setup and can be seen as a two-player minimax game for the networks to compete against each other.
A GAN uses two loss functions: one for generator training and one for discriminator training. 
The generator loss is calculated by how likely the generator is able to fool the discriminator, while the discriminator loss is calculated by how likely it is able to predict correctly. In other words, $D$ and $G$ try to find parameters to optimize the objective function $V(G, D)$:
\begin{equation}
    \underset{G}{\min} \text{ } \underset{D}{\max} \text{ } V(D, G) = \mathbb{E}_{x}[\log D(x)] + \mathbb{E}_{z}[\log(1 - D(G(z)))]. 
\end{equation}
Where $\mathbb{E}_{x}$ is the expected value over all real data instances, $D(x)$ is the discriminator's prediction if real data instance $x$ is real, $G(z)$ is the generator's output when given input $z$, $D(G(z))$ is the discriminator's prediction if a fake instance is real, and $\mathbb{E}_{z}$ is the expected value over all generated fake instances $G(z)$.

\subsection{SimGAN Overview}
With the explosion of deep learning models in recent years, it has become more important to collect as much data as possible.
However, this is not always possible, and learning from synthetic data may not achieve the desired performance due to a gap between synthetic and real data distributions.
To address the limitations in accessibility of real-world data or the poor assumptions and overly simplified simulations, researchers proposed a modification to the GAN framework named SimGAN \citep{SimGAN} that utilizes simulated images and unsupervised learning, such that synthetic images from a simulator are used as inputs instead of random vectors.
The model tries to improve the realism of the simulator’s output by trying to learn features from the real data, and then "refining" the simulated data to better match what is found in the real-world. Hence, the generative model is named the refiner ($R$).

As such, SimGANs have two new loss functions.
The refiner uses a self-regularization loss to minimize the difference between the simulated input and the resultant refined output, with the underlying idea being that the refined outputs should be a transformation of the original synthetic data and should not deviate too far from the original simulated conditions. The formula is defined below:   
\begin{equation}
    V(\theta) = - \underset{i}{\sum} \log(1 - D(R_{\theta}(x_{i}))) + \lambda || R_{\theta} (x_{i}) - x_{i}||_{1}.
\end{equation}
The first half of the equation is the same as the GAN generator loss.
In the second half, $\lambda$ is a regularization weight constant, $R_{\theta} (x_{i}) - x_{i}$ is the difference between the refiner output and the original input, and $||.||_{1}$ is the L1 norm.

The other loss is a local adversarial loss used by the discriminator network. 
Any local patch sampled from the refined image should have similar statistics to a real image patch.
Therefore, rather than defining a discriminator network that was trained on the entire global image, the authors used a discriminator that classifies all local image patches separately then aggregates them to predict whether the sample is real or fake.

Additionally, they introduced the idea of using a history buffer, where refined images generated by the previous steps of the refiner are sampled during each iteration of discriminator
training, and the discriminator loss function is computed by summing the predictions on the current batch of refined images and the sampled history of images. This is meant to improve the stability of adversarial training, as the discriminator network tends to only focus on the latest refined
images and the refiner network
may reintroduce previously seen artifacts that the discriminator has
forgotten about. 

\subsection{Easy Cartesian Genetic Programming (ezCGP)}
ezCGP is an end-to-end evolutionary Cartesian Genetic Programming framework designed to be highly flexible and customizable to any researcher’s task.
The graph representation of a genome, instead of a traditional tree representation, lends itself better to represent the graph architecture of a neural network. 
The framework also introduces the novel idea of compartmentalizing the genome into a sequence of ‘blocks’, each with their own evolutionary rules: a unique set of operators and hyperparameters, mating and mutation strategies, number of genes, and evaluation method. Each block can be thought as a conceptual component of an algorithm; by segmenting, we allow the algorithm to be evolved without mixing components from one concept to another. This allows for full representations of complex machine learning pipelines in evolutionary computation.
A more in-depth explanation of how it is applied in this work is included in Section \ref{Methods}. We share our code online for further research and experimentation.
 \footnote{\url{https://github.com/ezCGP/ezCGP}}

\subsection{Related Work}
Medical data and ECG synthesis is challenging since biological systems are dynamic with how the various parts of the body interact. Studies have attempted to use GANs to generate sufficiently realistic medical data \citep{TSTR} and one study has shown a successful example of using SimGANs to generate biologically plausible ECG signals\citep{SimGAN_ECG}.
The resultant synthetic data generated was shown to improve ECG classification when using the new synthetic data for training ECG classifiers.
In this example, they use the ECGSYN ECG simulator  \citep{advanced_ecg_sim} and tailor a specific self-regularization loss based on the simulator’s system of ordinary differential equations during the training process.
Unfortunately, the approach only generates examples for single heart beats, which are extremely short in nature and are not always useful when diagnosing heart disease. 

Integrating evolutionary techniques into traditional GAN approaches is not a novel idea.
Evolutionary GAN (E-GAN) \citep{E-GAN} evolves a population of generators, where the mutation method selects among different loss functions to train the generators, which then compete against a single discriminator. 
COEGAN \cite{COEGAN} utilizes neuroevolution to coevolve discriminator and generator architectures.
Multi-objective E-GAN (MO-EGAN) \citep{multi-EGAN} instead treats E-GAN training as a multi-objective optimization problem, and uses Pareto dominance, defined as when some other outcome is weakly preferred by all individuals and strictly preferred by at least one individual, to select the best solutions across the selected objectives that measure diversity and quality.
Other approaches propose improving GANs by discovering customized loss functions for each of its networks.
For example, TaylorGAN \citep{TaylorGAN} treats the GAN losses as Taylor expansions and optimizes custom definitions through multi-objective evolution. These losses were meant to act as an alternative to traditional GAN losses such as Wasserstein loss \citep{WGAN} or minimax loss. 
Lipizzaner \cite{Lipizzaner} uses spatial coevolution to evolve and train a grid of GANs simultaneously, and Mustangs \cite{Mustangs} builds upon E-GAN and Lipizzaner by mutating the loss function across the coevolution grid.

\section{Dataset}
\label{Dataset}

\subsection{ECG Dataset}
\begin{figure}[!htbp]
  \centering
  \includegraphics[trim=0 20 0 50, clip, width=\linewidth]{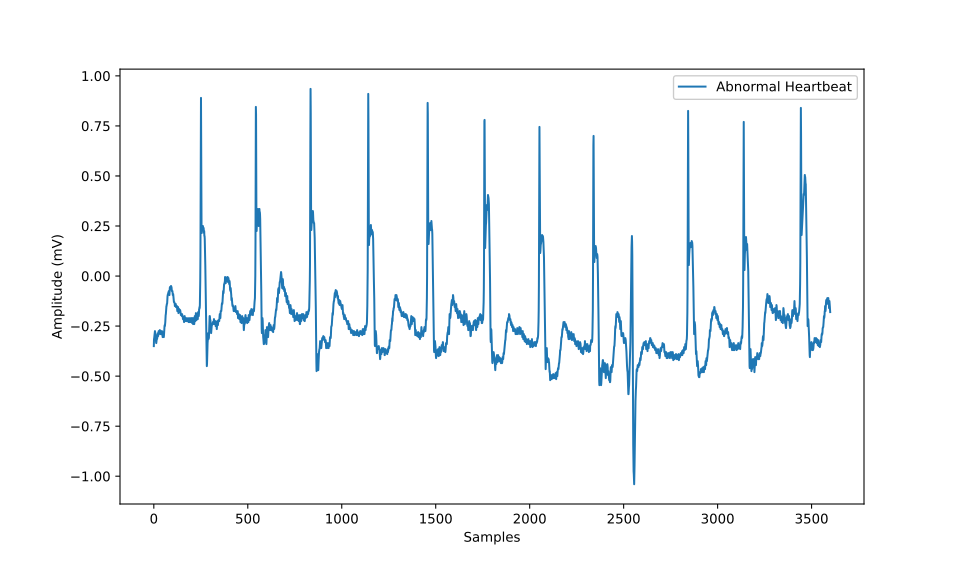}
  \caption{Example of an ECG heartbeat sample with an abnormal condition from the MIT-BIH Arrhythmia Database. Specifically, this sample contains an instance of a left bundle branch block beat around the 2500 mark}
  \Description{Abnormal heartbeat example}
  \label{abnormal_example}
\end{figure}

For a real-world example of 1D signals that are difficult to collect, we applied our approach to ECG heartbeat data to help generate realistic samples of various arrhythmias. In our approach to the problem, we want it to be generic towards the simulated source of data, i.e. it should work for multiple simulators, and we want our generated signals to be longer which better represent what doctors would actually use to diagnose heart disease.
We use ECG recordings taken from the MIT-BIH Arrhythmia Database  \citep{MIT-BIH} for real heartbeat data, as the database is a public well-established source of data for ECG heartbeat classification tasks.
The database contains 48 half-hour ECG records obtained from patients, where each record contains two 30-minute ECG lead signals collected with a sampling rate of 360 samples per second per channel. The database has annotations for heartbeat class information verified by independent experts.
However, a single heartbeat alone is not always useful for true heart disease classification, as such, longer recorded signals often prove more useful.
For this purpose, we have segmented the dataset samples into 10 second long intervals, where an entire segment is flagged as “abnormal” if any beats are not annotated as normal by the experts. In particular for training our SimGANs we selected 32 samples of uncommon abnormal signals, an example of which is shown in Figure \ref{abnormal_example}, but the whole dataset of both abnormal and normal heartbeats, roughly about 2000 and 5000 signals respectively, were later used for our ECG classifiers described in Section \ref{ECG_Classifier}.

\begin{figure}[!htbp]
  \centering
  \includegraphics[trim=0 20 0 50, clip, width=\linewidth]{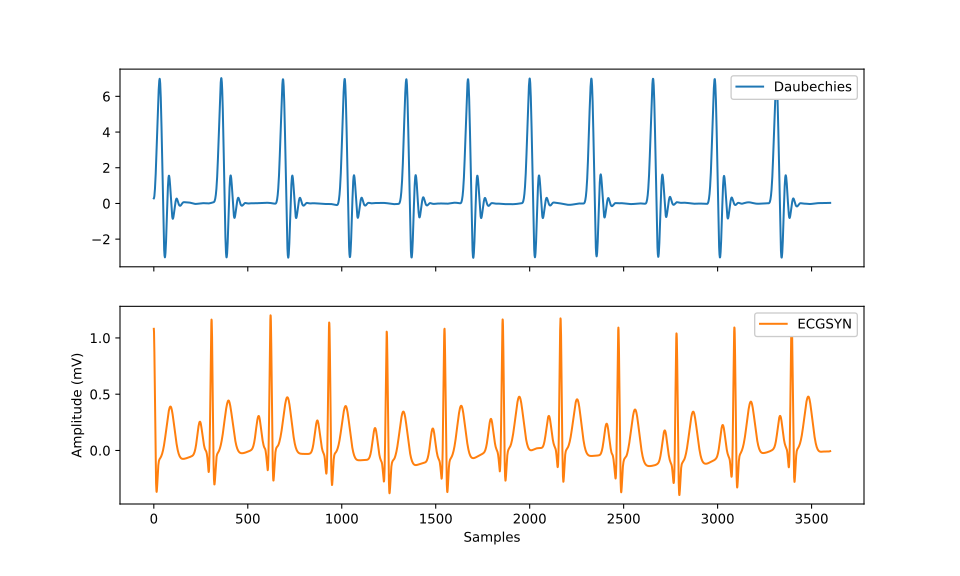}
  \caption{Generated ECG heartbeat samples from NeuroKit2, the top plot is using the Daubechies wavelets simulator and the bottom plot is using ECGSYN}
  \Description{Generated samples from NeuroKit2}
  \label{ECGSim_examples}
\end{figure}

For our simulated dataset, we use the NeuroKit2 \citep{NeuroKit2} package as our ECG simulator to collect samples.
NeuroKit2 is an open source Python package meant to provide easy access to advanced bio-signal processing routines based on existing cited works.
For our heartbeat ECG use case, NeuroKit2 can simulate healthy heartbeats and uses either a simple simulation based on Daubechies wavelets \citep{simple_ecg_sim}, which roughly approximate a cardiac cycle, or a more complex simulation based on ECGSYN \citep{advanced_ecg_sim}. 
We utilize both simulators and our simulated dataset contains an equal number of samples from both. 
Both simulators can specify conditions like duration, sampling rate, and heart rate, but ECGSYN can add synthetic noise drawn from a Laplacian distribution and even mimic a heart's random fluctuations between some beats.
To match our real ECG data from the MIT-BIH database, we use the same duration and sampling rate of 10 seconds and 360 samples per second respectively. Roughly 2500 samples were generated from both simulators with a range from 50 through 100 beats per minute for heart rate and up to 10 percent noise added for a total of 5000 samples. An example of a signal generated from both simulators is shown in Figure \ref{ECGSim_examples}. 

From these examples, we can see that the real samples often include noise from the collection sensors, and often are more complex and variable in the sequences of heartbeats.

\section{Research Methods}
\label{Methods}
SimGANs are notoriously difficult to train and evaluate, as during training we must balance the learning process between the refiner and discriminator. There is often no objective quantitative measure of how good a generated output is. Our training process is largely inspired by the original SimGAN framework. However, we propose new additions to the framework based on current research and apply the approaches to 1D data during training.

We add in specialized network architectures, generalizable feature-based quantitative evaluation metrics, and an evolutionary framework to help search and optimize potential neural architectures and hyperparameters. 

\subsection{Training Configuration}

We use the minimax loss described earlier as our adversarial loss between the refiner and discriminator, as just a simple binary cross entropy loss. For our self-regularization loss in our 1D space, it is actually simpler than the original, as we measure the absolute difference between two signals’ amplitudes at each point or bin. In terms of implementation, this is a simple L1 Loss.

The original SimGAN authors noticed that the refiner network tends to over-emphasize certain features to fool the discriminator networks, leading to the refiner network producing artifacts, so they proposed using a local adversarial loss. In our 1D use case, we quickly noticed that using just a local adversarial loss would lead to a collapse of the waveform and would not resemble the true wave structure, as signals and other forms of 1D data are often sequential and meaning is derived from ordering. Therefore, to help prevent the formation of artifacts, we propose a Siamese dual discriminator network where we ensemble a global discriminator network and a local discriminator network together to capitalize on both a global and local loss. The local discriminator network is fed segments of the signal, of specified equal length, then the network predicts if each segment is real or fake. The signal segment predictions are then averaged to obtain an aggregated prediction for the whole signal, which is then added to the prediction from the global discriminator network. Together they average their predictions to form the unified loss, which is backpropagated to update both network weights. Similarly, the local discriminator network also utilizes the history buffer and trains on previous iterations of refined waveforms.

Recent GAN approaches try to enforce a soft Lipschitz constraint using gradient penalties \citep{WGAN-GP}, to help bound the losses to some extent and help training stability, as the hypothesis is that mode collapse, when the refiner learns a single example that fools the discriminator and uses it over and over, is the result of the competing game converging to bad local equilibria. As such, during training, we chose to use DRAGAN \citep{DRAGAN} as our gradient penalty.

\subsection{Evaluation Metrics}
\label{sec:Evaluation}
These evaluation metrics are methods to measure a GAN's performance after it has completed training. Evaluating GANs is an open problem, as there is no objective truth in most cases. How does one classify if a generated signal is "real" enough? One way is simple visual qualitative analysis: "does it look right?" However, this often requires domain knowledge and is not scalable. The following metrics are a quantitative way to measure model performance. 

\subsubsection{Feature Evaluation}
One method for evaluating the refiner outputs is extracting key features from the real data, then comparing it to the features from the refined outputs. For every signal, we can effectively extract a feature vector. The following are a list of features we utilized, that can be generalized to multiple domains: 
\begin{enumerate}
    \item Amplitude/height difference between peaks 
    \item Amplitude/height ratio between peaks 
    \item Distance between peaks
    \item Trough heights
    \item Peak to trough amplitude/height difference
    \item Peak to trough distance
    \item Area under the signal's curve
    \item Number of times the signal changes direction
    \item Roughness of signal measured by a difference between a rolling mean of values and the current value
\end{enumerate}
Naturally, if the domain is deeply understood, more specific features can be used. We have ways to visualize the feature distributions so that we can qualitatively determine if a refiner truly transforms the simulated data to better reflect the real data distribution, or if one refiner is better than another based on one feature or another. 

\begin{figure}[h]
  \centering
  \includegraphics[trim=0 5 0 25, clip, scale=0.4]{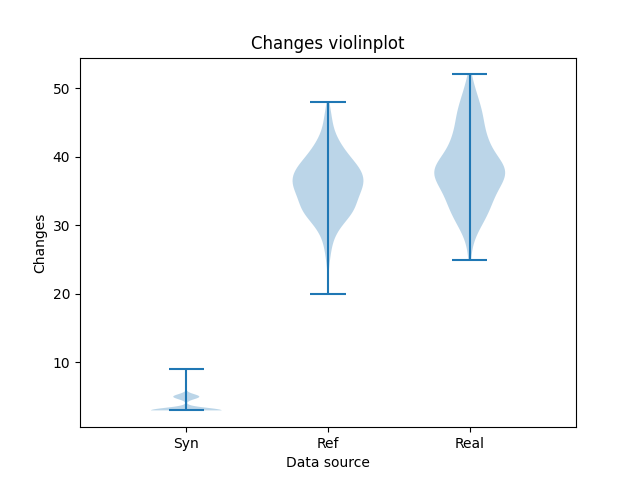}
  \caption{Example of a violin plot visualization of feature distributions of the number of times the signal changes direction. Note that we can qualitatively assess if the simulated distribution approaches the real distribution.}
  \Description{Violinplot example }
\end{figure}
We use distance measurements like Kullback–Leibler divergence (KL-Div), Wasserstein Distance, or Kolmogorov–Smirnov tests (KS-Stat) to quantitatively determine how far the refined output features are from our expected real features. We typically utilize KS-Stat as our distance metric, as it can be a nonparametric test of the equality between two discrete distributions. However, these distance metrics expect the distributions to be the same size and effectively "map" instances to one another. This is difficult for SimGANs as the simulated data size is expected to be much larger than the real data size, and there is no true mapping between real and simulated data. Therefore, to use distance metrics, we repeatedly sample the distributions until we are confident every signal has been sampled, and then average the distances calculated by our metric.

Another metric is a statistical t-test for unequal distributions to measure how likely that the two samples are drawn from the same distribution. We use a one-tailed unequal variance Welch’s t-test to compare the distributions of the extracted features from the refined and real signals. Unlike the traditional use of t-tests where we look for the p-value to be small to show there is a significant difference between two distributions, we want to show that there is no significant difference between the distribution of features between the real and refined data. Therefore, we want our p-value to be close to $1$. We can use the t-test on each feature individually, then keep track of the number of features where the simulated distribution was significantly different from the real distribution, and average the resultant p-values across all the features.

\subsubsection{FID Score}
Frechet Inception Distance \cite{fidscore} was originally used to measure the quality of generated images.
FID scores measure the difference between sets of generated and ground truth images via comparing the distributions (specifically the mean and co-variance) of model activations of each set.
An oracle network, typically Inception \cite{inception}, is used for computing the activations.
Unfortunately there is not enough data to train a custom oracle network for the ECG dataset. Therefore, we convert ECG waveforms to 2D images via stacking the waveform and zero-padding the width to the input shape of Inception. 

\subsubsection{Tournament Evaluation}
Tournament evaluation is a peer to peer evaluation method that is domain agnostic \citep{tournament}. The underlying idea is that we have our SimGAN population compete with themselves. So every refiner and discriminator will be paired with one another for every possible combination and the refiner is trying to fool the discriminator it is paired against. Each network is assigned an initial rating, based on the Glicko \citep{glicko} system, and as they win or lose matches, their rating will go up or down. The idea is that good refiners will be able to fool many types of discriminators if their outputs are close to the real signals.
However, as this method is peer-to-peer, it does not provide a direct measure of how well the refiners are performing and changes with the population. In fact, it is entirely possible that the entire population performs poorly. It should be noted that during our experiments, we noticed that the tournament would sometimes reward refiners that produced random noise, as the discriminators were not trained on random noise for very long. So despite a high tournament ranking, the resultant refiner is not guaranteed to generate good data. However, since GANs don't converge easily, a model at an earlier step can easily produce better outputs than ones in later steps. Therefore, we log each SimGAN at different intervals and use tournament evaluation to select the best point in time to represent the individual SimGAN as it is sent through ezCGP.

\subsection{Initial Seed}
\begin{figure*}[!htbp]
  \centering
  \includegraphics[width=.9\linewidth]{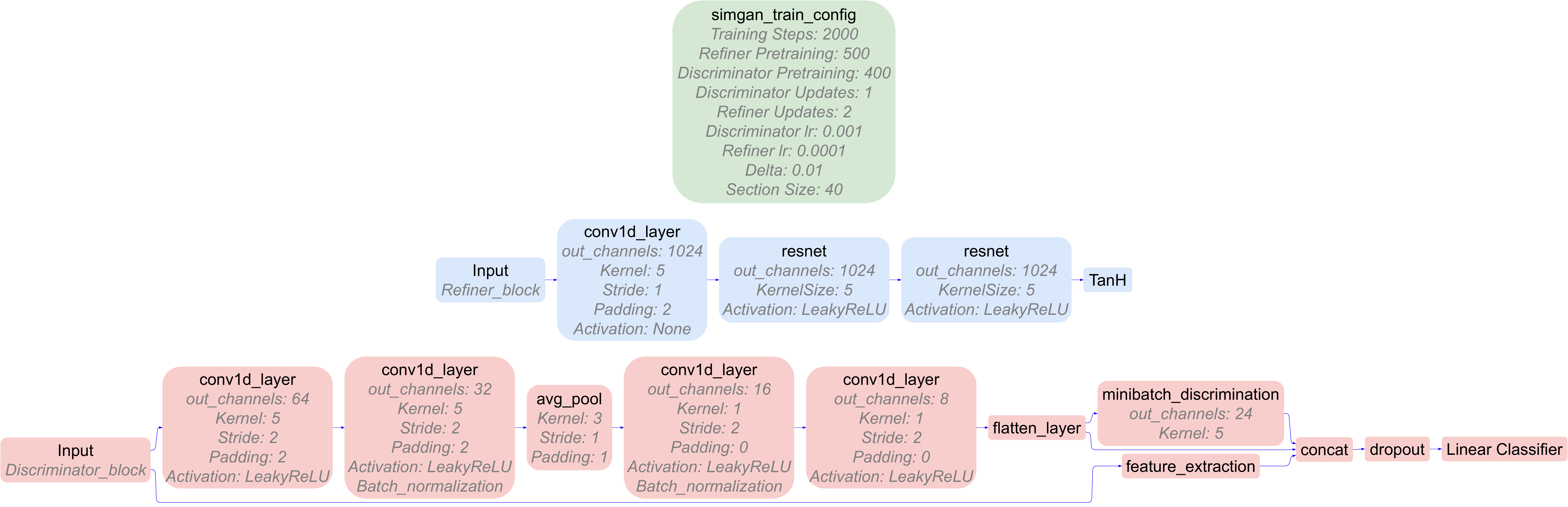}
  \caption{Architecture of our seeded individual for ezCGP. Notice the three distinct genome blocks. Note that only the active genes are shown.}
  \Description{seeded example }
  \label{seed}
\end{figure*}

With our new training improvements, we define a hand-designed SimGAN architecture that we hypothesized would be successful, and used it as an initial seed for our evolutionary process. Both the refiner and discriminator networks are inspired by DCGAN \citep{DCGAN} for deep convolutional network GANs where ResNet \citep{resnet} blocks are used as our convolutions. To further improve our discriminator so that it makes more accurate predictions while providing useful feedback to the refiner, we utilize the Siamese dual discriminator described earlier, a mini-batch discrimination \citep{mbd_feature_matching_loss} layer to discriminate between whole mini-batches of samples rather than between individual samples, and a feature extractor layer which extracts the features described above. This seed is visualized in Figure \ref{seed}.

\subsection{ezCGP Optimization}
Due to the compartmentalization of a genome in ezCGP, we define our genome as the following sequence of blocks: one block to build the architecture of the refiner network, another block to build the architecture of the discriminator network, and a third block to set the network training hyperparameters. Each block has its own set of operators, genes, and rule set to fit its definition. For our problem, we defined operators to act as various different neural network layer types, to better search for SimGAN improvements. A wide variety of operators and blocks allows for easy evolution and neural architecture search, especially if an initial seed is given. 
Example operators include: convolutional layers, concatenation layers, linear layers, pooling layers, dropout layers, batch normalization layers, ResNet blocks, activation layers, flatten layers, mini-batch discrimination layers, and custom feature extract layers.

Each operator has its own set of hyperparameters that can be tuned, but hyperparameters that guide the individual's learning process can also be tuned, such as optimizer types, learning rates, number of training steps, regularization weights, etc. All together these operators can easily allow us to build and define deep neural networks. Mating is defined to be strictly the exchange of whole blocks between parents; each block is set to mate with 33 percent probability. Mutation is defined as either a change of a gene's position in a block, a change of the operation used at a gene, or the change of the hyperparameter requested by the operation. The two neural network blocks have their genetic material mutated with a probability of 20 percent, and the remaining block mutated with a probability of 10 percent. After training, each individual was scored by the four objectives mentioned earlier: minimize the FID score, minimize the KS statistic, minimize the number of features where the simulated distributions were significantly different from the real distributions, maximize the average p-value of the feature distributions. The initial population size was set to 4, as a successful genome can easily spawn many children and a low initial population is used to prevent potential computational constraints such as GPU memory. We maintained a hall-of-fame with a max size of 40, and the next generation was selected using NSGA-II \citep{NSGA-II} from a pool of individuals: the previous generation’s offspring and hall-of-fame. Algorithm \ref{ezCGP_algo} shows the pseudo-code for the evolution process. We ran this for 48 hours on a single computer with 1 TeslaV100-PCIE 32GB GPU, 128 GB RAM, and a Xeon-Gold6126 CPU.


\begin{algorithm}
\caption{ezCGP Evolution}
\begin{algorithmic} 
\STATE $init\_population()$
\WHILE{not converged}
\STATE{$parents=get\_parent\_list()$}
\FOR{$p1,p2$ in $parents$}
\FOR{$i^{th}block$ in $range(num\_blocks)$}
\STATE{$roll=random\_number()$}
\IF{$roll < i^{th}block\_prob\_mate$}
\STATE{$population += mate(p1,p2)$}
\ENDIF
\ENDFOR
\ENDFOR

\FOR{$indiv$ in $population$}
\FOR{$i^{th}block$ in $range(num\_blocks)$}
\STATE{$roll=random\_number()$}
\IF{$roll < i^{th}block\_prob\_mutate$}
\STATE{$mutate(indiv)$}
\ENDIF
\ENDFOR
\ENDFOR

\FOR{$indiv$ in $population$}
\STATE{$train(indiv)$}
\STATE{$score(indiv)$}
\ENDFOR

\STATE{$population = select\_population(indiv)$}

\ENDWHILE

\end{algorithmic}
\label{ezCGP_algo}
\end{algorithm}

\section{Results and Discussion}
\label{ResultsAndDiscussion}

\subsection{Novel Configurations}
\begin{figure*}[!htbp]
  \centering
  \includegraphics[width=.9\linewidth]{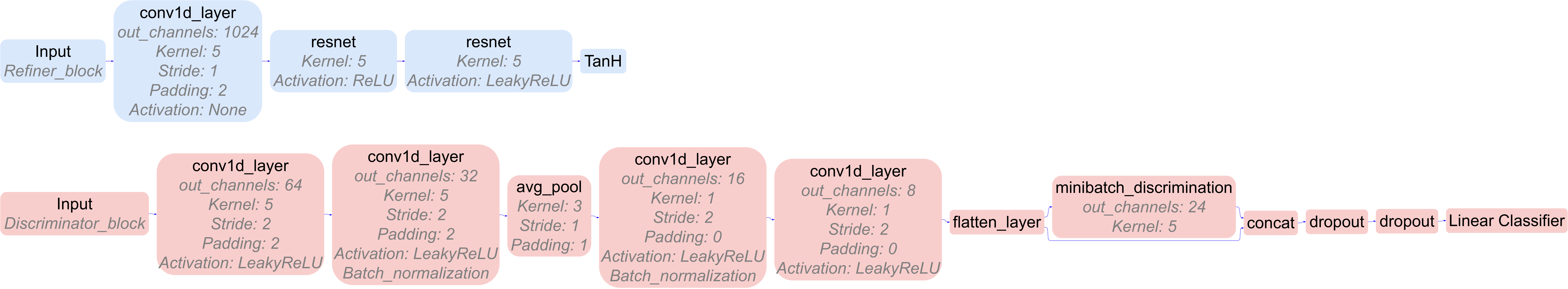}
  \caption{Architecture of an evolved individual from ezCGP. Notice the discriminator network opts not to use the feature extractor and adds additional dropout layers, causing a stronger regularization effect to occur. The refiner network uses a different activation function for the first ResNet block.}
  \Description{Evolved individual 1 example }
  \label{evolved1}
\end{figure*}

\begin{figure}[h]
  \centering
  \includegraphics[scale=0.5]{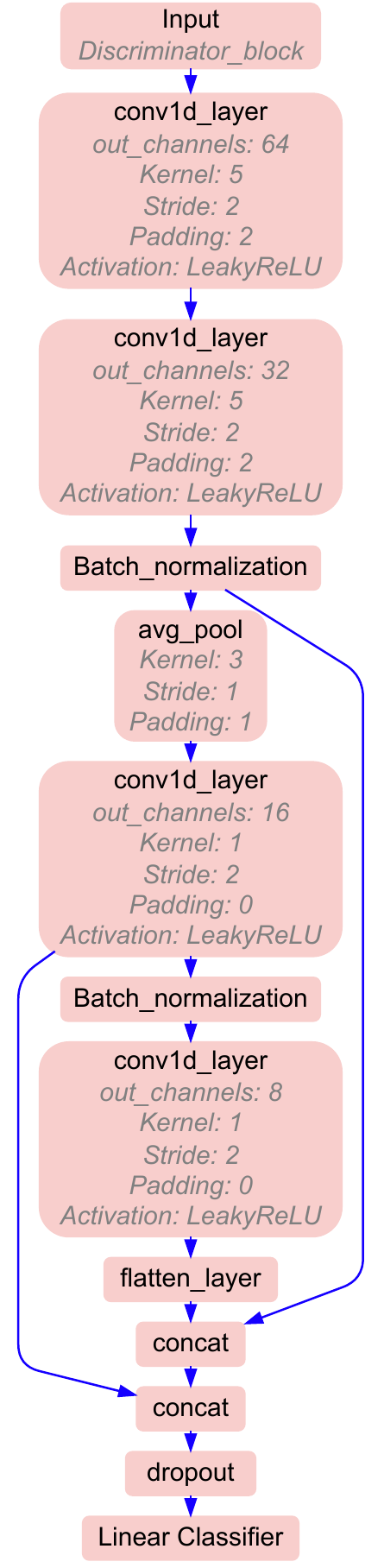}
  \caption{Architecture of another evolved individual's discriminator network from ezCGP. Notice the network takes advantage of concatenate layers to use features from previous convolutions.}
  \Description{Evolved individual 2 example }
  \label{evolved2}
\end{figure}

ezCGP searches for neural architectures, so over time we should see improved configurations for SimGANs across the population. Figure \ref{evolved1} and \ref{evolved2} are examples of evolved networks. In total over 40 individuals were created by the end of generation three. 

The evolved refiner genomes did not have as many significant changes as the discriminators. We hypothesize this is due to how we implemented a ResNet block as a large single gene-operator with less evolvability, and thus generally shows good performance overall and overpowers the discriminators. 


\subsection{Quality of Generated ECG Data}
\begin{figure}[!htbp]
  \centering
  \includegraphics[trim=0 65 0 65, clip, width=\linewidth]{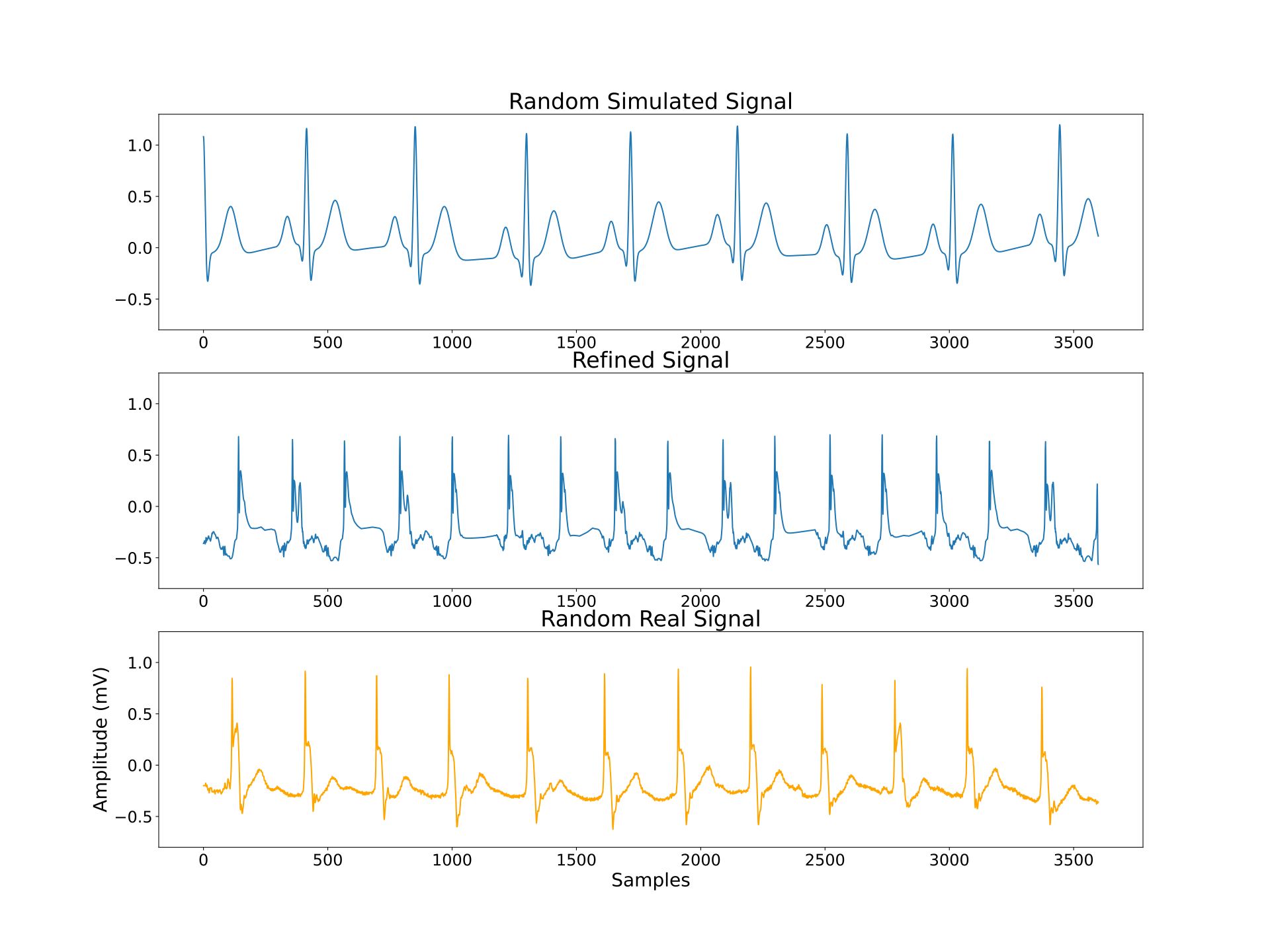}
  \caption{Generated ECG heartbeat sample from our seeded SimGAN individual in generation 0.}
  \Description{Generated sample from Gen 0 }
  \label{ECG_Gen_0}
\end{figure}

Figure \ref{ECGSim_examples} shows examples where we can see that the real samples for ECG heartbeat signals are often rougher than the simulated examples caused by noise from the collection sensors. Not only that, more complex conditions like blockages are not possible to generate with the given simulators. However, after training our seeded SimGAN model using the real and simulated data, we can qualitatively observe in Figure \ref{ECG_Gen_0} that the SimGAN's refined waveforms include noise and features present in the real data, though it may not be as principled as the simulation.

\begin{figure}[!htbp]
  \centering
  \includegraphics[trim=0 65 0 65, clip, width=\linewidth]{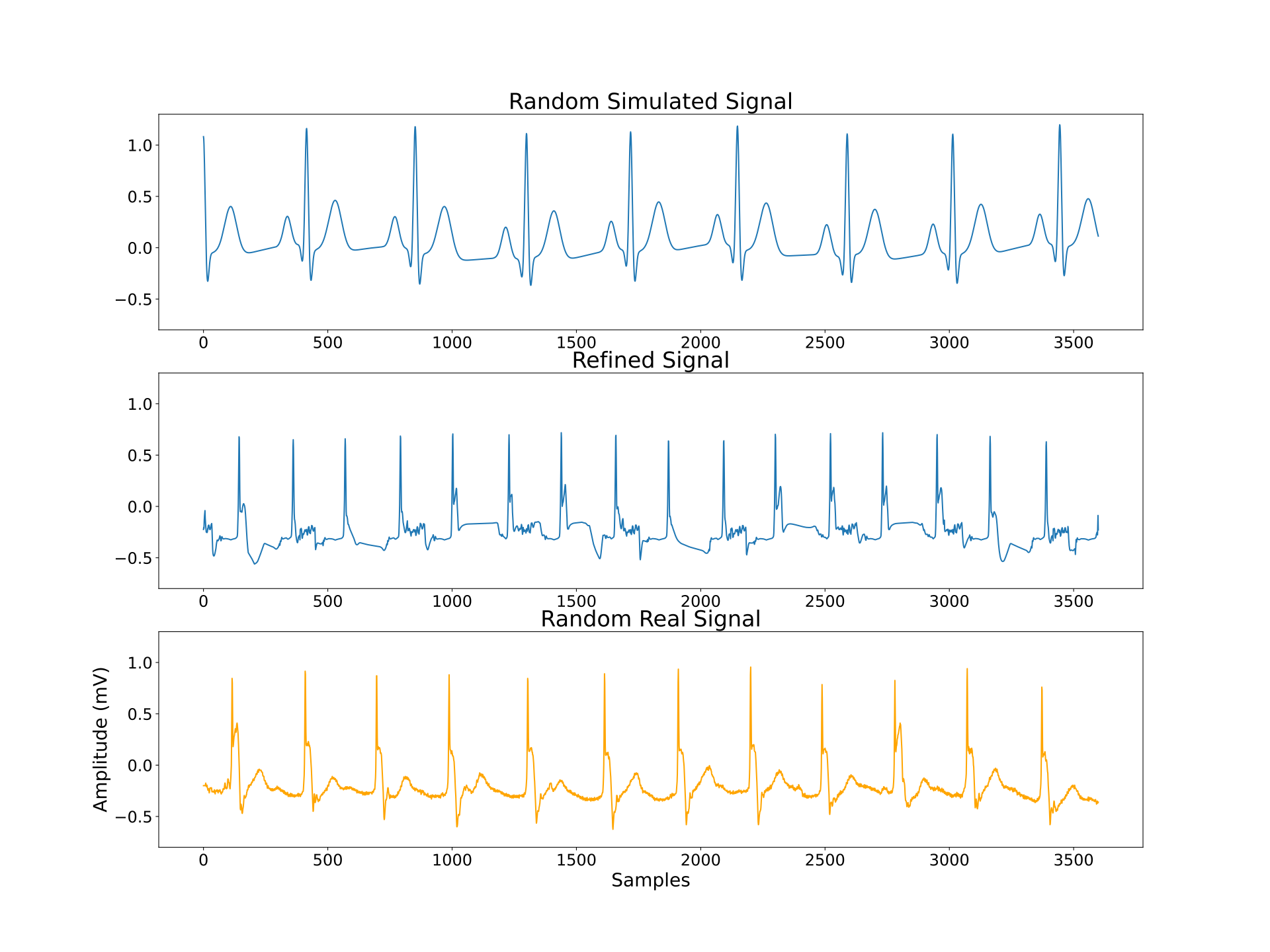}
  \caption{Generated ECG heartbeat sample from our optimized SimGAN individual from generation 9. }
  \Description{Generated sample from Gen 9}
  \label{ECG_Gen_5}
\end{figure}

To analyze the effects of evolution, we select a Pareto individual in our last generation from ezCGP and compare how the optimized refiner transforms the same wave. Figure \ref{ECG_Gen_5} shows improvement as the refined signal contains characteristics beyond learned noise.

\subsection{Empirical Analysis}
\label{ECG_Classifier}
Synthetic data is useful when it is sufficiently similar to real data. A TSTR score empirically shows that the data is similar enough to use, where the synthetic data is used to train a model, which is  tested on real data \citep{TSTR}. We extend TSTR score by training the same type of classifier on a subset of real data, a combination of real and simulated data, and a combination of real and refined data. We test the trained classifiers on the same withheld set of real signals. 

As Table \ref{Tab:comp} shows, we train four types of binary classifier models on the ECG signals output from our trained SimGAN models to predict whether a signal contains normal or abnormal characteristics. We can see that there is a performance boost in F1 scores after the models are trained on the refined data compared to just the real or real and simulated data. This indicates that the refined training data is likely closer to the real test set than the simulated training data was.  The refiner showcases how we can balance the training data with realistic abnormal heartbeats that were created from simulated healthy heartbeats. Similarly, the data our evolved SimGAN individual produces that the classifiers trained on sometimes performs better than the initial seeded individual, which suggests the evolved refiner's signals are of higher quality. Our refined signals are also model agnostic, as multiple models show similar improvement after being trained on the same data.


\begin{table}
\centering
\caption{F1-Scores of ECG classifiers with dataset variations}
\resizebox{\linewidth}{!}{%
\begin{tabular}{lrrrr}
\toprule
& AdaBoost
& Gradient Boosted Tree 
& Multi-Layer Perceptron
& 1D-SqueezeNet \citep{SqueezeNet}\\ 
\midrule
Real Waveforms 
& 0.205 & 0.256 & 0.293 & 0.875 \\ 
Real + Simulated Normal 
& 0.176 & 0.243 & 0.313 & 0.854 \\ 
Real + Seed-refined Abnormal  
& \textbf{0.247} & 0.341 & \textbf{0.491} & 0.884\\ 
Real + Evolved-refined Abnormal  
& 0.227 & \textbf{0.354} & 0.467 & \textbf{0.916}\\ \bottomrule
\end{tabular}%
}
\label{Tab:comp}
\end{table}

\section{Conclusion and Future Work}
\label{ConclusionAndFutureWork}
In this paper we show a method for using genetic evolution in conjunction with SimGANs to refine simulated data to better fit the real distribution.
We applied this method to an electrocardiogram dataset to generate refined waveforms, and show that using the refined waveforms to train an ECG classifying method will improve the classifier performance over the unrefined waveforms. 

We could potentially adapt this work to other forms of one-dimensional data, such as audio or electroencephalography data. Simulated data will help fill the datasets, but there is still a worry that simulated and real-world data are separable. The method could also be modified to work for unevenly-spaced time-series data. 

There is ample room for improvements in the evolution process. We can perform efficiency updates to encourage more diverse architectures to be found. Other seeds could be tested with various architectures that work well on one-dimensional data (such as transformers). Data preprocessing blocks and methods can be added as gene-operators, allowing ezCGP to specify individuals that both change model architecture/hyperparameters and the preprocessing for its inputs. Our dual discriminator network could be generalized for ensembled networks for both discriminator and refiner networks, and could lend itself to co-evolution solutions.

Given the unclear nature of evaluating GANs, multiple objectives are beneficial for evaluating them. However, too many objectives can make the selection process overly complex. While Section \ref{sec:Evaluation} showed a multitude of objectives for evaluation, we only used four for selection. Future work could explore creating more representative objectives or utilizing a many-objective selection algorithm.  

Since we have described a multitude of different losses and configurations which can be used to train SimGAN models, we can see that it quickly becomes very complex and there is no definitive reason for choosing one loss over another. There were  many other losses and training techniques that have shown success in other studies that we chose not to include, such as Wasserstein Loss \citep{WGAN} and the original WGAN-GP gradient penalty \citep{WGAN-GP}. Therefore, we propose to model the loss function as a symbolic regression problem, where all the losses and evaluations will be used, each weighted by an evolved constant, that can be optimized.


\bibliographystyle{ACM-Reference-Format}
\bibliography{sample-sigconf}


\appendix
\end{document}